\documentclass{bmvc2k}

\definecolor{cap_col}{rgb}{0,0,.4}
\usepackage{caption}
\usepackage{multirow}
\usepackage{wrapfig}
\usepackage{booktabs}
\usepackage{makecell}
\usepackage{listings}
\usepackage{balance}
\usepackage{placeins}
\usepackage{float}

\title{AR-TTA: A Simple Method for Real-World Continual Test-Time Adaptation}

\addauthor{Damian Sójka}{damian.sojka@doctorate.put.poznan.pl}{1,2}
\addauthor{Bartłomiej Twardowski}{btwardowski@cvc.uab.es}{1,3}
\addauthor{Tomasz Trzciński}{tomasz.trzcinski@pw.edu.pl}{1,4,5}
\addauthor{Sebastian Cygert}{sebcyg@multimed.org}{1,6}

\addinstitution{
 IDEAS-NCBR, Warsaw, Poland
}
\addinstitution{
 Institute of Robotics and Machine Intelligence \\
 Poznan University of Technology\\
 Poznań, Poland 
}
\addinstitution{
 Computer Vision Center \\
 Universitat Autònoma de Barcelona \\
 Barcelona, Spain
}
\addinstitution{
 Warsaw University of Technology\\
 Warsaw, Poland 
}
\addinstitution{
 Tooplox, Wroclaw, Poland 
}
\addinstitution{
 Gdańsk University of Technology\\
 Gdańsk, Poland 
}

\runninghead{Sójka et al.}{AR-TTA: A Simple Method for Real-World Continual TTA}

\def\eg{\emph{e.g}\bmvaOneDot}

\begin{document}

\maketitle

\begin{abstract}
   Test-time adaptation is a promising research direction that allows the source model to adapt itself to changes in data distribution without any supervision. 
   Yet, current methods are usually evaluated on benchmarks that are only a simplification of real-world scenarios. Hence, we propose to validate test-time adaptation methods using the recently introduced datasets for autonomous driving, namely CLAD-C and SHIFT.
   We observe that current test-time adaptation methods struggle to effectively handle varying degrees of domain shift, often resulting in degraded performance that falls below that of the source model.
   We noticed that the root of the problem lies in the inability to preserve the knowledge of the source model and adapt to dynamically changing, temporally correlated data streams.
   Therefore, we enhance the well-established self-training framework by incorporating a small memory buffer to increase model stability and at the same time perform dynamic adaptation based on the intensity of domain shift.
   The proposed method, named AR-TTA, outperforms existing approaches on both synthetic and more real-world benchmarks and shows robustness across a variety of TTA scenarios. 
   The code is available at \url{https://github.com/dmn-sjk/AR-TTA}.
\end{abstract}

\sloppy

\section{Introduction}

Deep neural networks have been shown to achieve remarkable performance in various tasks, however, they perform very well only when the test-time distribution is close to the training-time distribution. This poses a significant challenge since in a real-world application a domain shift can occur in many circumstances, \eg, weather change, time of day shift, or sensor degradation. For this reason, Test-Time Adaptation (TTA) methods have been widely developed in recent years~\cite{TENT,sun2020TTT}. They aim to adapt the source data pre-trained model to the current data distribution on-the-fly during test-time, using an unlabeled stream of test data.

An effective TTA method should work well regardless of a wide range of challenges, encompassing both abrupt and gradual domain shifts. 
Moreover, the model should be able to maintain stable performance even when handling lengthy sequences, potentially extending indefinitely.
Existing approaches are based on updating model parameters using pseudo-labels or entropy regularization~\cite{COTTA,TENT}. Further, filtering the less reliable samples is often employed to reduce error accumulation and improve the computational efficiency~\cite{eata,SAR,online_tta_eval}. However, all of the methods can become unstable due to the aforementioned difficulties, accumulate errors, make the pseudo-labels noisier, and cause performance degradation~\cite{chen2019accumulation}. Without using any source data the model is prone to \textit{catastrophic forgetting}~\cite{Mccloskey89} of previously acquired knowledge.

\begin{wrapfigure}{r}{0.445\textwidth}
  \begin{center}    
    \vspace{-2.5em}
    \includegraphics[width=0.45\columnwidth]{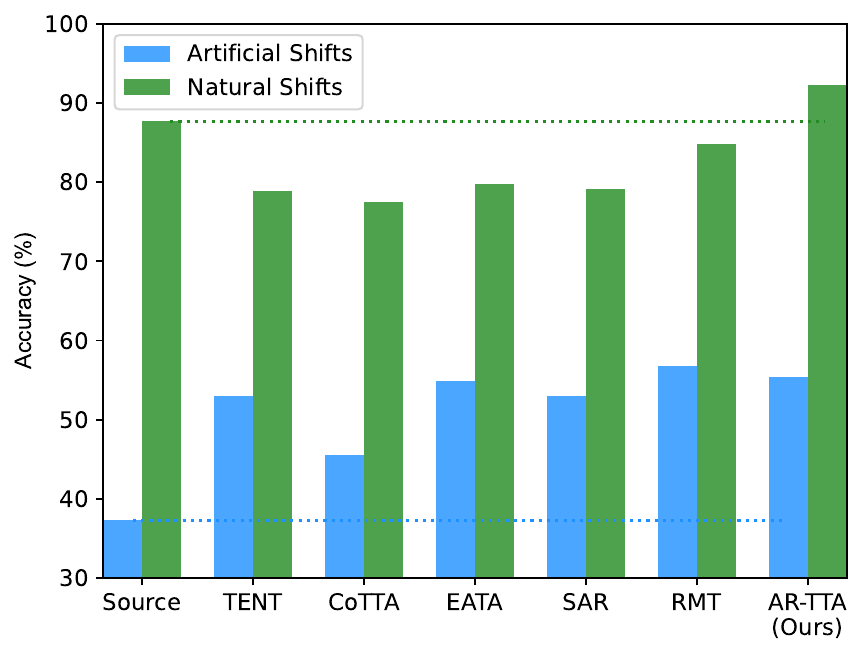}
  \end{center}
  \caption{Continual test-time adaptation methods evaluated on artificial (\mbox{CIFAR10C}, ImageNet-C~\cite{hendrycks2019robustness}) and natural (CIFAR10.1~\cite{cifar10_1}, SHIFT~\cite{shift2022}, \mbox{CLAD-C~\cite{verwimp2023clad}}) domain shifts. Our method is the only one that consistently allows to improve over the naive strategy of using the (frozen) source model.}
  \label{fig:teaser}
\end{wrapfigure}

Currently, methods are mostly evaluated on datasets with artificial, unrealistic domain shifts or relatively short-length sequences~\cite{TENT,COTTA,AdaContrast}, hence it is unknown how those methods will work in real-life scenarios with natural domain shifts. Therefore, we adapt the autonomous driving benchmark CLAD~\cite{verwimp2023clad}, to the continual adaptation setting. Moreover, we use a SHIFT dataset~\cite{shift2022}, which is synthetically generated by utilizing the realistic autonomous driving simulator CARLA~\cite{carla}, provides very long sequences, and allows us to specifically control for different domain factors (time of day, weather~conditions).

In the proposed evaluation setup, we find out that current approaches lack the required stability, as their performance significantly deteriorates compared to the source model (Fig.~\ref{fig:teaser}).
Additionally, we notice that they struggle to correctly estimate batch norm statistics with temporally correlated data streams and low batch sizes. 
In our TTA method for an image classification task, we extend popular self-training framework~\cite{COTTA} with a small memory buffer, which is used during adaptation to prevent knowledge forgetting, without relying on heuristic-based strategies or resetting model weights that are often used~\cite{COTTA,SAR}. Thanks to using mixup data augmentation~\cite{mixup}, a relatively small number of samples is required. Furthermore, we develop a module for dynamic batch norm statistics adaptation, which interpolates the calculated statistics of the pre-trained model and those obtained during testing, based on the intensity of domain shift.  
We call our method AR-TTA, as we improve \textbf{A}daptation by using dynamic batch norm statistics and maintain knowledge by \textbf{R}epeating samples from the memory buffer combined with mixup data augmentation.
As a result, our proposed method AR-TTA is simple, stable, and works well across a range of datasets with different shift intensities, when using small batches of data and over very long sequences.
Our main contributions can be summarized as follows:
\begin{itemize}
    \setlength\itemsep{0pt}
    \item We propose novel evaluation benchmarks for TTA based on autonomous-driving scenarios, which show significant limitations of existing TTA methods. 
    \item Based on discovered vulnerabilities, we propose a TTA method that dynamically updates the batch norm statistics based on
    the intensity of domain shift and use small memory buffer combined with mixup data augmentation. 
    \item Extensive evaluation shows that the proposed method obtains state-of-the-art performance on benchmarks with both artificial distortions and natural domain shifts.
\end{itemize}

\section{Related Work}

\noindent \textbf{Test-time adaptation (TTA).} 
Domain adaptation methods can be split into different categories based on what information is assumed to be available during adaptation~\cite{wang2018survey}. While in some scenarios, access to some labels in target distribution is available, the most common scenario is 
\textit{unsupervised domain adaptation}, in which the model has access to labeled source data and unlabeled target data at adaptation time. 
Additionally, in \textit{test-time adaptation} the model needs to adapt to the test-time distribution on-the-fly, in an \textit{online} fashion. In the test-time training (TTT) method~\cite{sun2020TTT} the model solves self-supervised tasks on the incoming batches of data to update its parameters.
TENT~\cite{TENT} updates only batch-norm weights to minimize predictions entropy. 
EATA~\cite{eata} further improves the efficiency of test-time adaptation methods, by using only diverse and reliable samples (with low prediction entropy). Additionally, it uses EWC~\cite{kirkpatrick2017ewc} regularization to prevent drastic changes in parameters important for the source domain.
Contrary to the TENT and EATA approaches, CoTTA~\cite{COTTA} updates the whole model. To prevent performance degradation it uses exponential weight averaging as well as stochastic model restoration, where randomly selected weights are reset to the source model.  SAR~\cite{SAR} further improves by removing noisy test samples with large gradients and using sharpness-aware minimization. Nevertheless, they also use model reset, to prevent forgetting. RMT~\cite{dobler2023CVPRmeanteacher} propose a mean teach framework and adapt it based on contrastive loss, self-training loss, and a simple replay strategy utilizing all of the source training data during test-time. Moreover, it assumes an additional warm-up step before the adaptation. Our method uses a small number of replay exemplars and does not require any extra steps before TTA.

\noindent\textbf{TTA benchmarks.}
The most popular setting for test-time adaptation includes applying different classes of synthetic corruptions proposed in~\cite{hendrycks2019robustness} to commonly used datasets, like CIFAR10~\cite{cifar} or ImageNet~\cite{imagenet} to artificially obtain domain shifts.
Another popular dataset for continual test-time adaptation is DomainNet~\cite{peng2019domainnet} which consists of images in different domains (e.g., sketches, infographics). Yet, the distribution shifts arising in real-world deployment, e.g., in autonomous driving, may be very dissimilar from different renditions of the same objects.
Hence, recently a CLAD, autonomous driving benchmark~\cite{verwimp2023clad} was introduced. It consists of naturally occurring distribution shifts like changes in weather and lighting conditions, traffic intensity, etc. It was developed for the supervised continual learning scenario. In this work, we use it for test-time adaptation, that is without using any label information.  
In our work, we also use SHIFT benchmark~\cite{shift2022}, a synthetically generated dataset for autonomous driving that captures the continuously evolving nature of the real world.
Similarly to us, CoTTA~\cite{COTTA} includes realistic domain shifts but their test is very small (1600 images).
To sum up, we extend over previous TTA work by focusing on realistic continual domain shifts over very long sequences.

\section{Method}

\begin{figure}[t!]
    \centering
    \includegraphics[width=0.62\columnwidth]{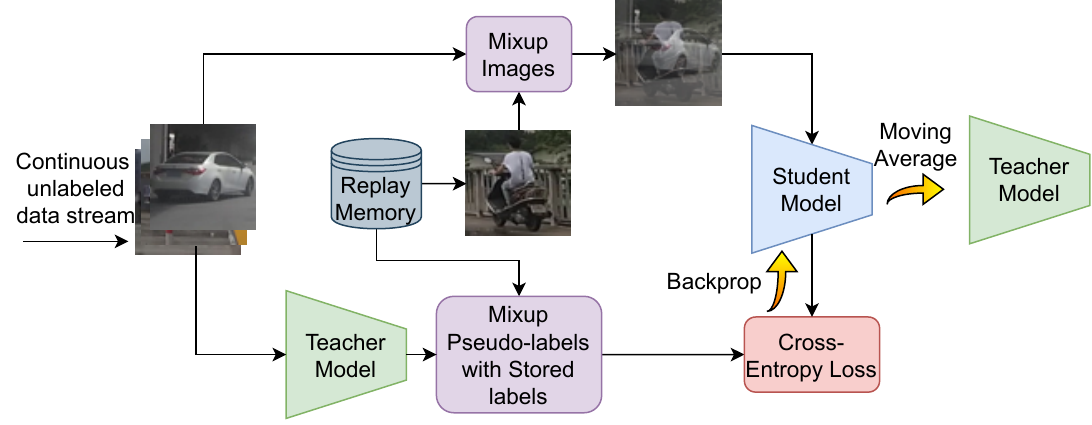}
    \vspace{2mm}
    \caption{
    Our method, AR-TTA, utilizes a replay strategy and the mean teacher framework. 
    Each image is paired with an exemplar sampled from memory and image pairs are mixed up. Similarly, pseudo-labels from the teacher model are mixed up with the labels of sampled exemplars. The student model is updated based on cross-entropy loss between its predictions on augmented samples and augmented pseudo-labels.
    The teacher model is adapted based on an exponential moving average of student's weights. Predictions for each image are taken from the teacher model.}
    \label{fig:method_diagram}
\end{figure}

Continual TTA aims to adapt the pre-trained model $f_{\theta_0}$ trained on the labeled source data $(\mathcal X^S, \mathcal Y^S)$ to the ever-changing stream of unlabeled test data batches $\textbf{x}^T$ on-the-fly during the evaluation.
Our proposed approach (Fig.~\ref{fig:method_diagram}.) can be divided into three parts. We start the description by introducing the model update procedure in Section~\ref{sec:emateacher}. Then, in Section~\ref{sec:mixup} we explain the usage of experience replay with mixup augmentation. The process of adapting batch normalization statistics is presented in Section~\ref{sec:bn_stats} 

\subsection{Weight-averaged Consistency} \label{sec:emateacher}
Following previous works~\cite{dobler2023CVPRmeanteacher, COTTA}, we use the self-training mean teacher framework to update the model. It improves the reliability of pseudo-labels and minimizes the risk of error accumulation.
We initialize two identical artificial neural network models, student model $f_\theta$ and teacher model $f_{\theta^\prime}$, with the equal weights obtained by training on source data. For each batch of test data $\textbf{x}_t^T$ at time step $t$, predictions are derived from both models. Teacher model predictions $\hat{y}_{t}^{\prime T}$ are used as soft pseudo-labels. The student model is updated by the cross-entropy loss between its predictions and the pseudo-labels:

\begin{equation}
\label{eq:student_update}
    \mathcal{L}_{\theta_t}(\textbf{x}_t^T) = -\sum_{c} \hat{y}_{t, c}^{\prime T} \log\hat{y}_{t, c}^T
\end{equation}
where $\hat{y}_{t, c}^T$ is the probability of class $c$ predicted by the student model. 
Next, teacher's weights $\theta^{\prime}$ are updated by exponential moving average of student's weights $\theta$:
\begin{equation}
\label{eq:teacher_update}
    \theta_{t+1}^{\prime} = \alpha\theta_t^{\prime} + (1 - \alpha)\theta_{t+1}
\end{equation}
where $\alpha$ is a smoothing factor. 
Using a weight-averaged teacher model ensures less noisy pseudo-labels~\cite{COTTA}, improves generatlization~\cite{ema_analysis}, and added inertia prevents rapid weights update based on noisy self-training feedback.

We do not limit the weights update only to affine parameters of batch normalization layers, as in many other TTA methods~\cite{eata, TENT, SAR}, and we adapt the whole model. We argue that adapting only batch normalization layer weights does not give the model enough flexibility to perform successfully on varying domains. We confirm this claim experimentally in the appendix.

\subsection{Experience Replay with Adaptation} \label{sec:mixup}

During continual test-time adaptation, self-training feedback is not guaranteed to be accurate, and frequent model updates inevitably strive for significant error accumulation, catastrophic forgetting, and even model collapse~\cite{SAR}. 
Continual learning works show that experience replay is one of the most effective strategies for mitigating catastrophic forgetting~\cite{icarl, hal, xder, verwimp2023clad}. Following this insight we propose to use the replay strategy during adaptation to remind the model what it has learned and strengthen its initial knowledge.
Building on this, we integrate Mixup data augmentation~\cite{mixup} to enhance model robustness, inspired by works~such~as~\cite{lump, dualaug}.

After completing the pre-training of the source model, we store a low, equal for each class, number of random exemplars from the labeled source data in the memory. 
In each test-time adaptation iteration, we randomly sample exemplars $\textbf{x}_t^{S}$, along with their labels $\textbf{y}_t^S$, from memory. The number of sampled exemplars is equal to the batch size. Mixed up batch of samples $\Tilde{\textbf{x}}_t$ is generated by linearly interpolating samples from test data with samples~from~memory:
\begin{equation}
\label{eq:mixup_samples}
    \Tilde{\textbf{x}}_t = \lambda \textbf{x}_t^{T} + (1 - \lambda) \textbf{x}_t^{S}
\end{equation}
where $\lambda \sim$ Beta$(\psi, \rho)$, for $\psi, \rho \in (0,\infty)$.
Similarly, labels for cross-entropy loss are the result of interpolation between pseudo-labels produced by the teacher model based on the current unmodified test batch $\hat{y}_{t}^{\prime T}$ and labels $\textbf{y}_t^S$ from the memory, with the same $\lambda$:
\begin{equation}
\label{eq:mixup_labels}
    \Tilde{\textbf{y}}_t = \lambda \hat{\textbf{y}}_{t}^{\prime T} + (1 - \lambda) \textbf{y}_t^{S}
\end{equation}
Student model takes augmented batch $\Tilde{\textbf{x}}_t$ as input. Its predictions are compared with interpolated labels $\Tilde{\textbf{y}}_t$ to calculate the loss as described in the previous Section~\ref{sec:emateacher}.

A similar approach to mixing exemplars from replay memory with the ones to train on was successfully used in LUMP~\cite{lump}, however, they used this method for the continual representation learning task.
Using experience replay along with the Mixup augmentation helps the model preserve already obtained knowledge and makes the noisiness of pseudo-labels less impactful for the adaptation process.

\subsection{Dynamic Batch Norm Statistics} \label{sec:bn_stats}
Batch normalization~\cite{batch_norm} (BN) layers normalize network values using running statistics from initial training data, applied during test-time. However, out-of-distribution data can disrupt this process, leading to poor model performance. Test-time adaptation methods~\cite{COTTA, eata, TENT, SAR} often recalculate statistics per each batch separately. However, this approach can be flawed due to small sample sizes and temporal data correlation. In these cases, using BN statistics from source data may offer a more accurate distribution estimation.

To perform a robust statistics estimation, we take the inspiration from~\cite{mecta} and propose to calculate BN statistics $\phi_t = (\mu_t, \sigma_t)$ at time step $t$ during test-time by linearly interpolating between saved statistics from source data $\phi^S$ and calculated values from current batch $\phi^T_t$:
\begin{equation}
\label{eq:bn_interpolation}
    \phi_t = (1 - \beta_{ema})\phi^S + \beta_{ema} \phi^T_t 
\end{equation}
where $\beta_{ema}$ is a parameter that weights the influence of saved and currently calculated statistics. Since the severity of the distribution shift might vary, we need to adequately adjust the value of $\beta_{ema}$. It should be large in cases when the distribution shift is severe compared to the source domain and low when the distributions are similar. Following~\cite{mecta}, we utilize the symmetric KL divergence as a measure of distance between distributions $D(\phi_{t-1}, \phi_t^T)$:
\begin{equation}
\label{eq:sym_kl_div}
    D(\phi_{t-1}, \phi_t^T) = \frac{1}{C}\sum_{i=1}^{C}KL(\phi_{t-1, i} || \phi_{t, i}^T) + KL(\phi_{t, i}^T || \phi_{t-1, i})
\end{equation}
The distance is used to calculate $\beta_t$ at time step $t$:
\begin{equation}
\label{eq:new_beta}
    \beta_t = 1 - e^{-\gamma D(\phi_{t-1}, \phi_t^T)}
\end{equation}
where $\gamma$ is a scale hyperparameter. To compensate for the fact that the current distribution can be wrongly estimated and to provide more stability for the adaptation, we take into account previous $\beta_{1:t-1}$ values and use an exponential moving average for $\beta_{ema}$ using $\alpha$ coefficient:
\begin{equation}
\label{eq:beta_ema}
    \beta_{ema} = (1 - \alpha)\beta_{t-1} + \alpha \beta_t
\end{equation}

Our method differs from MECTA~\cite{mecta} by preserving the original BN statistics instead of using exponential moving averages. We are motivated by the fact that changing this value causes the inevitable forgetting,
and susceptibility to temporal correlation. We ensure consistent performance on similar domains to the source data.
At the same time, we allow for a slight drift away from source data statistics, by using the exponential moving average of $\beta$ parameter, giving enough flexibility for the adaptation to severe domain shifts.

\section{Experiments}
\noindent \textbf{Datasets and Benchmarks.}
We evaluate the methods on multiple image classification datasets with domain shifts. We use CIFAR10C and ImageNet-C as benchmarks with artificial, corruption-based shifts, widely utilized for evaluation of the test-time adaptation~approaches~\cite{SAR, COTTA,eata, dobler2023CVPRmeanteacher, TENT}. 
The experiments with natural shifts include tests on CIFAR10.1~\cite{cifar10.1} and SHIFT~\cite{shift2022} datasets, and CLAD-C continual learning benchmark~\cite{verwimp2023clad} adapted to the test-time adaptation setting. 

CIFAR10.1 was designed to minimize the distribution shift relative to the original CIFAR10 dataset. This allows us to test the robustness of TTA methods to very delicate distribution shifts.
The goal of CLAD-C benchmark was to introduce a more realistic testing bed for continual learning. 
The images taken at different locations and conditions are chronologically ordered, inducing distribution shifts in labels and domains. It consists of 6 test sequences. We pre-train the source model on the first train sequence. TTA is continually tested on the 5 remaining ones with a total number of 17092 images.
In the SHIFT dataset, images are taken in numerous types of realistic domains simulated in a virtual environment, including different weather and times of day. We consider it a natural domain shift benchmark due to its realistic representation of distribution shifts, which are induced by real conditions rather than image corruptions. 
We train the source model on images taken in clear weather during the day and test the adaptation methods for various weather combinations and times of day. We end up with 14 different domains resulting in a total of 380667 images. 
We called our original continual test-time adaptation benchmark the SHIFT-C.
More details can be found in the appendix.

\noindent \textbf{Methodology.}
We examine the continual test-time adaptation setup where the model is continually adapted to new domains without resetting the model unless it is part of a tested method. To simulate the continuous stream of data and the need for the model to adapt quickly, we use a low batch size of 10. This is also the batch size commonly used in the online continual learning~\cite{online_continual}. In embedded applications, limited resources often mean smaller batch sizes, necessitating TTA methods to work effectively with fewer samples.
We assess the methods using mean classification accuracy and average mean class accuracy (AMCA). The latter calculates the mean accuracy over all classes averaged for each domain, making every class equally important in class-imbalanced datasets.

\noindent \textbf{Baselines.}
We conduct experiments involving six state-of-the-art methods as baselines: TENT-continual~\cite{TENT}, EATA~\cite{eata}, CoTTA~\cite{COTTA}, SAR~\cite{SAR}, and RMT~\cite{dobler2023CVPRmeanteacher}. Moreover, we show results for discarding BN statistics from source data and calculating the statistics for each batch separately (BN-1)~\cite{bn_stats_adapt}. We also report the performance of the frozen source model (Source). To provide a fair comparison, the results without a replay strategy of RMT and our method (AR-TTA) are presented.

\noindent \textbf{Implementation Details.}
Following other state-of-the-art TTA works, we use pre-trained WideResnet28/ResNet50 models from \textit{RobustBench}~\cite{croce2020robustbench} model zoo for CIFAR10C and CIFAR10.1/ImageNet-C datasets. On the rest of the benchmarks, we utilize ResNet50 architecture with weights pre-trained on ImageNet obtained from \textit{torchvision} library~\cite{torchvision2016} and finetuned to the source data for the specific benchmark.

Our method utilizes the default replay memory size of 2000 samples, which is commonly used in continual learning settings~\cite{masana2022class} and adds only minor storage requirements. The RMT method utilizes all of the source
training data, following the original work. The rest of the details regarding the technical side of experiments can be found in the appendix.

\subsection{Results} \label{sec:results}

The results for all of the tested benchmarks are presented in Table~\ref{tab:all}. Our method can achieve the best average accuracy of 75.5\% indicating the robustness of our approach. Moreover, it shows a solid performance even without the replay method.

\noindent \textbf{Artificial Domain Shifts.}
Artificial domain shifts pose a great challenge for source models, achieving only 56.5\%/17.1\% mean accuracy for CIFAR10C/ImageNet-C. Calculating BN statistics for each batch separately, already significantly improves the result to 75.0\%/26.9\% accuracy on corrupted images. Each of the compared state-of-the-art TTA methods uses the BN-1 technique, therefore their performance improves over it, but the increase in accuracy value is not that significant. RMT outperformed our method on CIFAR10C, however, it lacks performance on other benchmarks and utilizes the whole source training dataset for replay, rendering AR-TTA a more reliable method for different kinds of datasets.

\begin{table}[h!]
\centering
\caption{Classification accuracy~(\%) for all of the tested online continual test-time adaptation tasks. Methods that use exemplars are in the right section. Red color indicates accuracy lower compared to a source model.}
\vspace{2mm}
\label{tab:all}
\scalebox{0.75}{
    \begin{tabular}{l|ccccc|c}
        \hline
        Method & CIFAR10C & ImageNet-C & CIFAR10.1 & CLAD-C & SHIFT-C & Average\\
        \hline
    
        Source & 56.5 & 18.1 & 88.3 & 81.3 & 93.5 & 67.5 \\
        
        BN-1 & 75.0 & 26.9 & \textcolor{red}{81.3} & \textcolor{red}{71.1} & \textcolor{red}{85.1} & 67.9 \\
        
        TENT~\cite{TENT} & 76.7 & 29.2 & \textcolor{red}{82.3} & \textcolor{red}{71.5} & \textcolor{red}{82.7} & 68.5 \\
        
        EATA~\cite{eata} & 78.2 & 31.5 & \textcolor{red}{82.9} & \textcolor{red}{71.1} & \textcolor{red}{85.1} & 69.8 \\
        
        CoTTA~\cite{COTTA} & 75.7 & 15.5 & \textcolor{red}{82.3} & \textcolor{red}{72.6} & \textcolor{red}{77.4} & \textcolor{red}{64.7} \\
        
        SAR~\cite{SAR} & 75.2 & 30.8 & \textcolor{red}{81.3} & \textcolor{red}{71.1} & \textcolor{red}{85.1} & 68.7 \\
        
        RMT w/o replay~\cite{dobler2023CVPRmeanteacher} & 77.6 & 21.7 & \textcolor{red}{80.6} & \textcolor{red}{75.1} & \textcolor{red}{92.2} & 69.4 \\
        
        RMT~\cite{dobler2023CVPRmeanteacher} & \textbf{83.1} & 30.5 & \textcolor{red}{83.3} & \textcolor{red}{75.3} & \textbf{95.9} & 73.6 \\

        \hline
        
        AR-TTA (Ours) w/o replay & 77.3$_{\pm0.07}$ & 30.0$_{\pm0.45}$ & 88.2$_{\pm0.10}$ & \textbf{83.9}$_{\pm0.30}$ & \textcolor{red}{92.4}$_{\pm0.25}$ & 74.4 \\
        AR-TTA (Ours) & \textbf{78.8}$_{\pm0.13}$  & 32.0$_{\pm0.07}$ & \textbf{88.3}$_{\pm0.05}$ & 83.7$_{\pm0.64}$ & 94.8$_{\pm0.03}$ & \textbf{75.5} \\
        
        \hline
    \end{tabular}}
\end{table}

\noindent \textbf{Natural Domain Shifts.}
The BN-1 method significantly degrades the performance of the frozen source model on natural shifts by about 8-10 percentage points of accuracy. Similarly, the state-of-the-art TTA methods achieve significantly lower mean accuracy than the Source, rendering them ineffective for natural domain shifts. It suggests that benchmarking such methods on artificial domain shifts in the form of corruptions is not a reliable estimate of the TTA method's performance in practical applications. 
Our method, which uses pre-calculated statistics and exemplars of source data during adaptation, outperformed state-of-the-art methods and achieved higher accuracy than the source model. This indicates the effectiveness and adapting capabilities. Keeping the pre-calculated statistics intact might sometimes be more beneficial, especially on domain shifts on which the source model performs relatively well.

\begin{figure}[h]
    \centering
    \includegraphics[width=0.70\columnwidth]{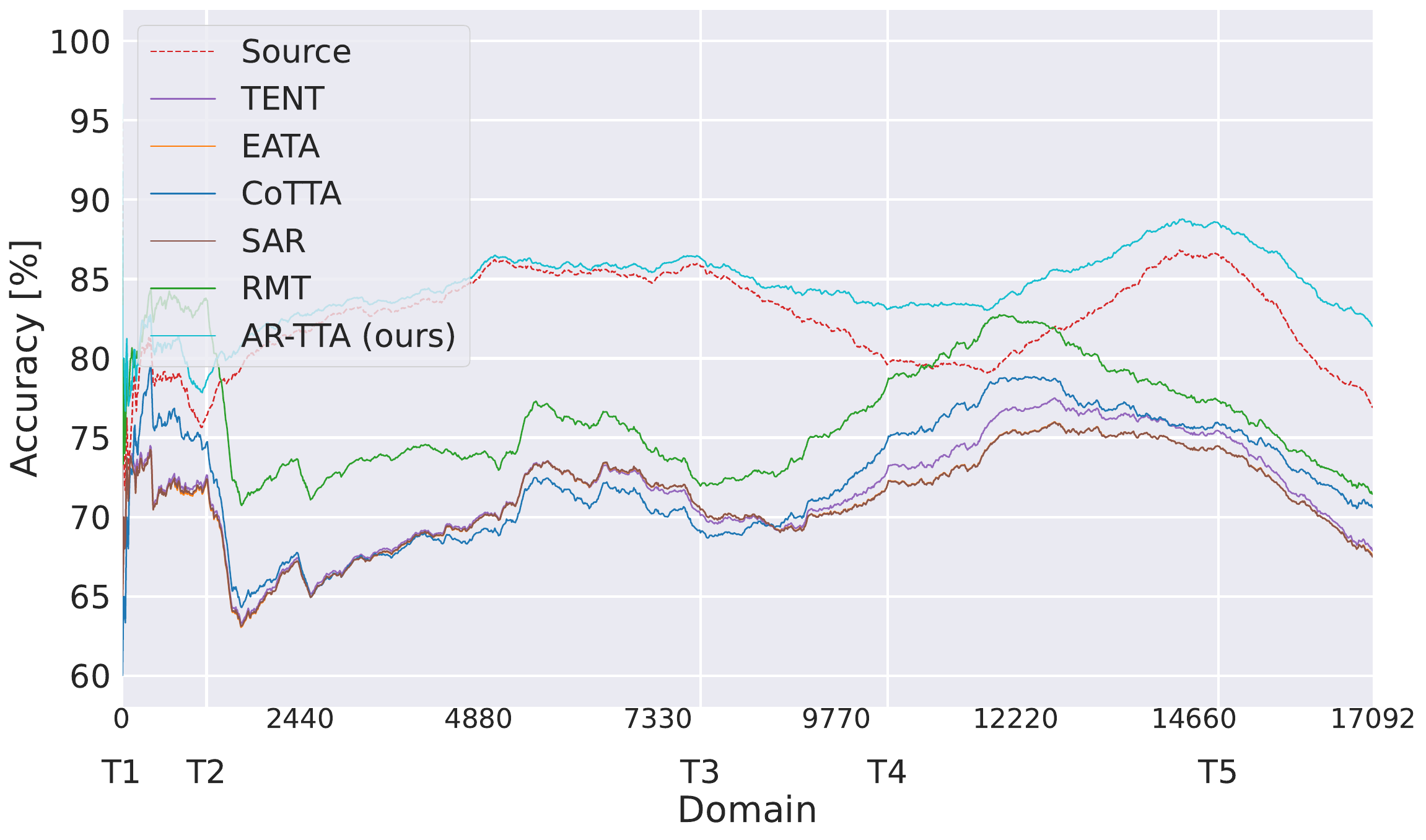}
    \vspace{2mm}
    \caption{Batch-wise classification accuracy (\%) averaged in a window of 400 batches on CLAD-C benchmark for the chosen methods continually adapted to the sequence of data, with major ticks on the x-axis symbolizing the beginning of a different domain and minor ticks indicating image number. Best viewed in color.}
    \label{fig:clad_batchwise}
    \vspace{-1.em}
\end{figure}

\noindent \textbf{Accuracy Over Time.}
Results in Figure~\ref{fig:clad_batchwise} show how the accuracy of state-of-the-art methods drops below the source model performance. We can see that AR-TTA maintains the accuracy of the source model on the first two domains where the source model works well,  and then improves over it when the source model performance deteriorates (at domain T3). 

\begin{table*}[h!]
\centering
\caption{Classification accuracy and average mean class accuracy (AMCA)~(\%) for the CLAD-C continual test-time adaptation task.}
\vspace{2mm}
\label{tab:cladc}
\scalebox{0.7}{
\begin{tabular}{l|ccccc|cc|cc}
\multicolumn{1}{l}{}& \multicolumn{5}{c}{$t\xrightarrow{\hspace*{3cm}}$}& \\ \hline
Method & T1 & T2 & T3 & T4 & T5 & Mean day & Mean night & Mean & AMCA \\ \hline

Source & 75.6 & 85.9 & 73.3 & 87.5 & 66.2 & 86.6 & 71.2 & 81.3 & 57.6 \\

BN-1 & 73.2 & 69.9 & 75.0 & 75.5 & 59.7 & 72.2 & 69.1 & 71.1 & 48.3 \\

TENT~\cite{TENT} & 73.4 & 69.8 & 76.5 & 76.1 & 59.7 & 72.4 & 69.8 & 71.5 & 47.6 \\

EATA~\cite{eata} & 73.3 & 69.9 & 75.0 & 75.6 & 59.7 & 72.2 & 69.1 & 71.1 & 48.4 \\

CoTTA~\cite{COTTA} & 75.2 & 69.3 & 80.2 & 77.0 & 62.7 & 72.4 & 72.9 & 72.6 & 44.8 \\

SAR~\cite{SAR} & 73.2 & 69.9 & 75.0 & 75.5 & 59.7 & 72.2 & 69.1 & 71.1 & 48.3 \\

RMT w/o replay~\cite{dobler2023CVPRmeanteacher} & \textbf{87.1} & 70.9 & \textbf{86.6} & 76.9 & 64.3 & 73.9 & \textbf{79.3} & 75.1 & 48.4 \\

RMT~\cite{dobler2023CVPRmeanteacher} & 83.8 & 71.3 & 85.0 & 77.6 & 66.4 & 74.5 & 78.4 & 75.3 & 48.8 \\

\hline

AR-TTA (Ours) w/o replay & 76.9 & \textbf{86.7} & 81.4 & 87.9 & \textbf{73.5} & 87.2 & 77.1 & \textbf{83.9}$_{\pm0.30}$ & 59.6$_{\pm2.92}$ \\
AR-TTA (Ours) & 77.2 & \textbf{86.7} & 80.0 & \textbf{89.6} & 70.7 & \textbf{87.8} & 75.7 & 83.7$_{\pm0.64}$ & \textbf{63.1}$_{\pm3.32}$ \\

\hline
\end{tabular}}
\end{table*}

\begin{table*}[ht!]
\centering
\caption{Classification accuracy and average mean class accuracy (AMCA)~(\%) for the SHIFT-C continual test-time adaptation task.}\label{tab:shift}
\vspace{2mm}
\scalebox{0.57}{
\begin{tabular}{l|cccc|ccccc|ccccc|cc}
\multicolumn{1}{l}{}& \multicolumn{14}{c}{ $t\xrightarrow{\hspace*{14cm}}$}& \\
\hline
\multirow[b]{2}{*}{Method} &
\multicolumn{4}{c|}{daytime} & \multicolumn{5}{c|}{dawn/dusk} & \multicolumn{5}{c|}{night} & 
\multirow[b]{2}{*}{Mean} & \multirow[b]{2}{*}{AMCA} \\

& \rotatebox[origin=c]{70}{cloudy} & \rotatebox[origin=c]{70}{overcast} & \rotatebox[origin=c]{70}{rainy} & \rotatebox[origin=c]{70}{foggy} & 
\rotatebox[origin=c]{70}{clear} & \rotatebox[origin=c]{70}{cloudy} & \rotatebox[origin=c]{70}{overcast} & \rotatebox[origin=c]{70}{rainy} & \rotatebox[origin=c]{70}{foggy} & 
\rotatebox[origin=c]{70}{clear} & \rotatebox[origin=c]{70}{cloudy} & \rotatebox[origin=c]{70}{overcast} & \rotatebox[origin=c]{70}{rainy} & \rotatebox[origin=c]{70}{foggy} \\ \hline

Source & \textbf{97.9} & \textbf{98.2} & \textbf{97.5} & 92.5 & 93.6 & 94.1 & 94.0 & 93.5 & 91.5 & 89.1 & 89.3 & 90.6 & 89.1 & 90.7 & 93.5 & 89.5 \\

BN-1 & 89.1 & 88.9 & 88.0 & 86.2 & 85.3 & 84.8 & 87.3 & 83.5 & 84.8 & 81.3 & 81.2 & 80.3 & 79.6 & 83.5 & 85.1 & 69.9 \\

TENT~\cite{TENT} & 89.6 & 88.8 & 87.5 & 84.6 & 83.3 & 81.2 & 85.0 & 80.7 & 80.2 & 78.0 & 77.0 & 76.1 & 75.7 & 77.6 & 82.7 & 57.6 \\

EATA~\cite{eata} & 89.1 & 88.9 & 88.0 & 86.2 & 85.3 & 84.8 & 87.4 & 83.6 & 84.9 & 81.4 & 81.4 & 80.3 & 79.7 & 83.7 & 85.1 & 70.5 \\

CoTTA~\cite{COTTA} & 88.2 & 87.1 & 84.1 & 80.5 & 78.7 & 76.2 & 80.5 & 74.0 & 74.9 & 71.5 & 70.3 & 67.3 & 64.9 & 66.2 & 77.4 & 47.2 \\

SAR~\cite{SAR} & 89.1 & 88.9 & 88.0 & 86.2 & 85.3 & 84.8 & 87.3 & 83.5 & 84.8 & 81.3 & 81.2 & 80.3 & 79.6 & 83.6 & 85.1 & 69.9 \\


RMT w/o replay~\cite{dobler2023CVPRmeanteacher} & 93.0 & 94.2 & 93.6 & 91.7 & 91.0 & 91.3 & 93.0 & 92.2 & 90.8 & 90.7 & 90.9 & 91.7 & 90.5 & 91.7 & 92.0 & 80.3 \\
RMT~\cite{dobler2023CVPRmeanteacher} & 95.9 & 97.2 & 97.0 & \textbf{95.4} & \textbf{95.4} & \textbf{95.9} & \textbf{96.5} & \textbf{96.3} & \textbf{95.0} & \textbf{94.9} & \textbf{95.1} & \textbf{96.3} & \textbf{95.5} & \textbf{95.6} & \textbf{95.9} & \textbf{91.1} \\

\hline

AR-TTA (Ours) w/o replay & 96.4 & 96.5 & 95.3 & 93.2 & 92.2 & 91.9 & 93.2 & 91.4 & 91.8 & 88.7 & 88.7 & 88.6 & 87.5 & 91.2 & 92.4$_{\pm0.25}$ & 83.5$_{\pm0.96}$ \\

AR-TTA (Ours) & 97.7 & 98.0 & 97.4 & 94.3 & 94.2 & 95.5 & 94.8 & 95.2 & 93.1 & 92.3 & 92.7 & 93.0 & 91.4 & 92.6 & 94.8$_{\pm0.03}$ & 90.2$_{\pm0.24}$ \\

\hline
\end{tabular}}
\end{table*}


\begin{table}[t!]
\captionsetup{skip=0.6\baselineskip} 
\parbox{.49\linewidth}{
\vspace{-2.5mm}
\centering
\caption{Classification accuracy~(\%) for CIFAR10C and CLAD-C tasks for different configurations of the proposed method.}
\vspace{4.5mm}
\scalebox{0.7}{

    \begin{tabular}{l|c|c}
    \hline
    Method & CIFAR10C & CLAD-C \\
    \hline
    \textbf{A}: Weight-avg. teacher & 75.7$_{\pm0.07}$ & 71.1$_{\pm0.53}$ \\
    \textbf{B}: A + Replay memory & 77.3$_{\pm0.16}$ & 69.0$_{\pm0.66}$ \\
    \textbf{C}: B + Mixup & 78.5$_{\pm0.13}$ & 72.2$_{\pm0.31}$ \\
    \hline
    \textbf{D}: A + Dynamic BN stats & 77.3$_{\pm0.07}$ & 83.8$_{\pm0.82}$ \\
    \textbf{E}: D + Replay memory & 79.8$_{\pm0.03}$ & 82.8$_{\pm1.09}$ \\
    \hline
    \textbf{AR-TTA (Ours)}: E + Mixup & 78.8$_{\pm0.13}$ & 83.7$_{\pm0.64}$ \\
    \hline
    \end{tabular}}
\label{tab:each_component}
}
\hfill
\parbox{.49\linewidth}{
\centering
\caption{Classification accuracy and average mean class accuracy (AMCA)~(\%) results for state-of-the-art methods with simple \textbf{replay method added}. 
}
\scalebox{0.7}{
\begin{tabular}{l|c|cc}
\hline

\multirow{2}{*}{Method} & CIFAR10C & \multicolumn{2}{c}{CLAD-C} \\

& Mean & Mean & AMCA \\ 

\hline

Source &
56.5 & 81.3 & 57.6 \\

\hline

TENT~\cite{TENT} &
77.3 & 70.3 & 49.2 \\

EATA~\cite{eata} &
78.6 & 71.1 & 48.4 \\

CoTTA~\cite{COTTA} &
\textbf{79.9} & 72.6 & 51.0 \\

SAR~\cite{SAR} &
75.3 & 71.1 & 48.3 \\

\hline
AR-TTA (Ours) &
78.8 & \textbf{83.7} & \textbf{63.1} \\
\hline
\end{tabular}}

\label{tab:baselines_w_replay}
}
\end{table}

\subsection{Detailed Analysis} \label{sec:ablation}

\noindent \textbf{Effect Of Exemplars.}
Tables~\ref{tab:cladc} and~\ref{tab:shift} present the AMCA results for CLAD-C and SHIFT-C benchmarks. 
The results show a significant improvement in mean per-class accuracy while using a replay method. This suggests that the usage of replay memory might be important for keeping the performance for each class high in class-imbalanced TTA setups. Also, on average, our method is still better than that of our competitors when all methods have used exemplars and is the only one that consistently improves over the frozen model.

\noindent \textbf{Component Analysis.}
Table~\ref{tab:each_component} shows the contribution of individual components used in the proposed method. For the initial setup \textbf{A}, we used a weight-averaged teacher model to generate pseudo-labels and cross-entropy loss to adapt the model of which the BN statistics from source data are discarded. Adding a simple replay method (\textbf{B}, \textbf{E}) by injecting randomly augmented exemplars from memory to the batch in a 1:1 ratio, did not improve the performance on every dataset. It can be seen that mixup data augmentation can boost the performance of a simple replay method (\textbf{C}). Moreover, dynamic BN statistics significantly contribute to the accuracy increase (\textbf{D}, \textbf{E}, \textbf{AR-TTA}), especially on the CLAD-C benchmark.

\noindent \textbf{Baselines With Simple Replay Memory.}
To ensure a fair comparison and check the state-of-the-art approaches' performance with the use of a replay strategy, we added a simple replay method to each of them. A constant number of replay exemplars, equal to the batch size, was sampled on each batch from a class-balanced replay memory. The value of cross entropy loss calculated from exemplars was added to the original loss of each method. The results are in the Table~\ref{tab:baselines_w_replay}. While the proposed method performs slightly worse than CoTTA on the CIFAR10C, it performs significantly better on the natural domain shift dataset. Most importantly, our method is the only one that constantly improves over the source model.

\noindent \textbf{Computational Efficiency Evaluation}
Table~\ref{tab:time_comparison} presents a comparison of wall-clock time and memory usage for baseline methods and our approach on 10,000 images from CIFAR10C benchmark. While our method does not rank as the most computationally efficient, it achieves a balance between computational demands and performance. 
Notably, despite incorporating exemplars, we maintain a consistent computational budget. 
This is because the additional samples are utilized within the mix up augmentation, leaving the training batch size unchanged. This approach contrasts with the RMT method, where the use of replay techniques substantially increases computational complexity.

\begin{table}[h!]
\centering
\caption{The wall-clock time (seconds) and memory usage (MB) measured for processing 10,000 images of \mbox{CIFAR10C} on a single RTX 4080 GPU.}
\vspace{2mm}
\label{tab:time_comparison}
\scalebox{0.75}{
    \begin{tabular}{l|cc}
        \hline
        Method & Time [s] & Memory [MB]\\ 
        \hline
    
        Source & 8.0 & 304 \\
        
        BN-1 & 8.3 & 304 \\
        
        TENT~\cite{TENT} & 16.3 & 506 \\
        
        EATA~\cite{eata} & 24.3 & 505 \\ 
        
        CoTTA~\cite{COTTA} & 319.4 & 1532 \\ 
        
        SAR~\cite{SAR} & 30.8 & 506 \\
        
        RMT~\cite{dobler2023CVPRmeanteacher} w/o replay& 55.5 & 1576 \\
        
        RMT~\cite{dobler2023CVPRmeanteacher} & 163.7 & 3039 \\
        \hline
        
        AR-TTA (Ours) w/o replay & 66.2 & 1098 \\ 
        AR-TTA (Ours) & 66.6 & 1098 \\ 
        
        \hline
    \end{tabular}}
\end{table}

\vspace{-2.em}
\section{Conclusion}

In this paper, we evaluate existing continual test-time adaptation (TTA) methods in real-life scenarios using more realistic data by proposing two new evaluation benchmarks, namely SHIFT-C and CLAD-C. 
Our findings reveal that current state-of-the-art methods are inadequate in such settings, as they fall short of achieving accuracies better than the frozen source model. 
This raises concerns about the applicability of certain TTA methods in the real world and sheds light on the frequent model resets observed in some approaches.
To address these limitations, we propose a novel and straightforward method called AR-TTA, based on the self-training framework. AR-TTA utilizes a small memory buffer of source data, combined with mixup data augmentation, and dynamically updates the batch norm statistics based on the intensity of domain shift. 

Through experimental studies, we demonstrate that the AR-TTA method achieves state-of-the-art performance on various benchmarks. 
These benchmarks include realistic evaluations with small batch sizes, long test sequences, varying levels of domain shift, as well as artificial scenarios such as corrupted CIFAR10-C. 
Notably, AR-TTA consistently outperforms the source model, which serves as the ultimate baseline for feasible TTA methods.
Our more realistic evaluation of TTA with a variety of different datasets provides a better understanding of their potential benefits and shortcomings.

\noindent\textbf{Limitations.} 
The main limitation of our method is that we use a memory buffer from the source data, which might be an issue in resource-constrained scenarios or if there are some~privacy~concerns.   

\section*{Acknowledgments}
This research was partially funded by National Science Centre, Poland, grant no: 2020/39/B/ST6/01511, 2022/45/B/ST6/02817 and 2023/51/D/ST6/02846.
Bartłomiej Twardowski acknowledges the grant RYC2021-032765-I.
We gratefully acknowledge Polish high-performance computing infrastructure PLGrid (HPC Center: ACK Cyfronet AGH) for providing computer facilities and support within computational grant no. PLG/2023/016613. 
This paper has been supported by the Horizon Europe Programme (HORIZON-CL4-2022-HUMAN-02) under the project "ELIAS: European Lighthouse of AI for Sustainability", GA no. 101120237.

\bibliography{egbib}

\newpage

\appendix
\section{Appendix}

\renewcommand{\thefigure}{A.\arabic{figure}}
\setcounter{figure}{0}
\renewcommand{\thetable}{A.\arabic{table}}
\setcounter{table}{0}

\sloppy

Section~\ref{sec1} provides a detailed analysis of our method: effect of memory replay, adapting different model layers, and hyper-parameter analysis.
Section~\ref{sec2} provides detailed results of the experiments from the main paper. Section~\ref{sec3} described implementation details including the hyper-parameter search for all of the methods. Section~\ref{sec4} described benchmarks used in this paper, including the introduced ones SHIFT-C and CLAD-C.

\subsection{AR-TTA detailed analysis}
\label{sec1}

\subsubsection{Effect Of Replay Memory Size}

\begin{figure}[h!]
    \centering
    \includegraphics[width=0.7\columnwidth]{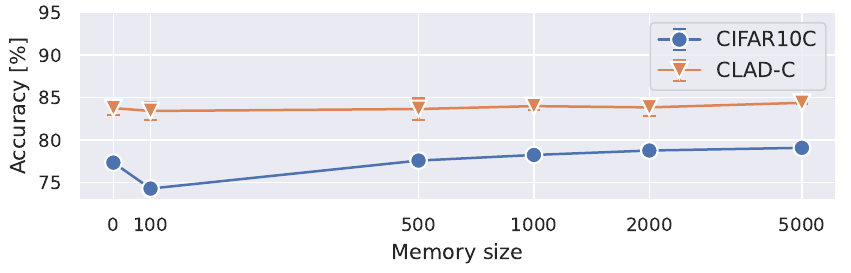}
    \vspace{1mm}
    \caption{The influence of replay memory size on the resulting accuracy on CIFAR10C and CLAD-C benchmarks.}
    \label{fig:mem_size}
    \vspace{-5mm}
\end{figure}

The necessity to keep a set of samples from source data in memory can be problematic in memory-limited settings. We verified the possibility of minimizing the size of replay memory and evaluated our method with different numbers of stored samples. The results in Figure~\ref{fig:mem_size} show that our method is robust to replay memory size. There is no significant difference in accuracy between memory sizes of 500 and 10000 for both CIFAR10C and CLAD-C benchmarks. A slight degradation in performance can be seen with only 100 exemplars for CIFAR10C. Less severe domain shift in CLAD-C allows for a more significant reduction in the memory size without the performance drop.

\subsubsection{Adapted Weights}
Table~\ref{tab:adapted_weights} shows the performance related to different configurations of adapted weights with our proposed method. We check the multiple configurations of adapting the last two layers, the first two layers, and only BN statistics. The best results are achieved by adapting the whole model.

\begin{table*}[h!]
\centering
\caption{Classification accuracy~(\%) for different configurations of adapted weights with our proposed method AR-TTA.}
\vspace{2mm}
\label{tab:adapted_weights}
\scalebox{0.75}{
\begin{tabular}{l|c||l|c}
\hline
\multicolumn{2}{c}{CIFAR10C (WideResNet28)} & \multicolumn{2}{c}{CLAD-C (ResNet50)} \\
\hline
Adapted weights & Mean & Adapted weights & Mean \\
\hline

block 1 (BN affine only) & 61.4 & layer 1 (BN affine only) & 81.7 \\
block 1, 2 (BN affine only) & 26.3 & layer 1, 2 (BN affine only) & 81.5 \\
block 2, 3 (BN affine only) & 73.8 & layer 3, 4 (BN affine only) & 81.9 \\
block 3 (BN affine only) & 72.8 & layer 4 (BN affine only) & 82.1 \\
\hline

block 1 & 17.4 & layer 1 & 80.4 \\
block 1, 2 & 13.9 & layer 1, 2 & 81.6 \\
block 2, 3 & 78.4 & layer 3, 4 & 83.3 \\
block 3 & 77.2 & layer 4 & 82.8 \\

\hline
BN affine only & 75.1 & BN affine only & 81.8 \\
\hline
Whole model (Ours) & \textbf{78.8} & Whole model (Ours) & \textbf{83.7} \\

\hline

\end{tabular}
}
\end{table*}

\subsubsection{Influence Of Beta Distribution Shape For Mixup Augmentation}

\begin{wraptable}{r}{0.5\textwidth}
\vspace{-4mm}
\centering
\caption{Classification accuracy~(\%) for \mbox{CIFAR10C} and CLAD-C tasks for different configurations of beta distribution parameters $\psi$ and $\rho$ for sampling interpolation parameter $\lambda \sim$ Beta$(\psi, \rho)$ required for mixup data augmentation.}
\label{tab:beta_distribution}
\vspace{2mm}
\scalebox{0.8}{
\begin{tabular}{cc|cc}
\hline

$\psi$ & $\rho$ & CIFAR10C & CLAD-C \\ \hline
5.0 & 5.0 & 78.6 & 83.7 \\
1.0 & 5.0 & 78.6 & 81.8 \\
5.0 & 1.0 & 78.2 & 83.0 \\
2.0 & 8.0 & 74.8 & 82.0 \\
8.0 & 2.0 & 77.5 & 83.8 \\
\hline
\multicolumn{4}{c}{Ours} \\
\hline
0.4 & 0.4 & \textbf{78.8} & \textbf{83.7} \\

\hline

\end{tabular}}
\vspace{-8mm}
\end{wraptable}

The beta distribution in mixup augmentation is used to sample interpolation parameter between exemplars. Within our method, it controls the interpolation between test data samples and exemplars from the replay memory bank. By shaping this distribution we can adjust what are the fractions of replay and test data in the augmented samples. Results are shown in Table~\ref{tab:beta_distribution}. The shape of the distribution did not have a significant impact on the results. The symmetric shape of the distribution, common for mixup augmentation, gives the best results.   

\subsubsection{Additional Component Analysis}
Table~\ref{tab:each_component_pseudolabels} shows results for different component configurations of our method. It includes the experiment without the usage of a weight-averaged teacher. We utilized pseudo-labels from the adapted model itself (configuration \textbf{A}). Additionally, we show the performance of our method when chosen exemplars for replay memory are not class-balanced. 

\begin{table}[h!]
\centering
\caption{Classification accuracy~(\%) for CIFAR10C and CLAD-C tasks for different configurations of the proposed method.}
\label{tab:each_component_pseudolabels}
\vspace{2mm}
\scalebox{0.76}{
\begin{tabular}{l|c|c}
\hline

Method & CIFAR10C & CLAD-C \\

\hline

\textbf{A}: Pseudo-labels & 75.5$_{\pm0.07}$ & 71.3$_{\pm0.54}$ \\
\textbf{B}: A + Weight-avg. teacher & 75.7$_{\pm0.07}$ & 71.1$_{\pm0.53}$ \\

\textbf{C}: B + Replay memory & 77.3$_{\pm0.16}$ & 69.0$_{\pm0.66}$ \\
\textbf{D}: C + Mixup & 78.5$_{\pm0.13}$ & 72.2$_{\pm0.31}$ \\

\hline

\textbf{E}: B + Dynamic BN stats & 77.3$_{\pm0.07}$ & 83.8$_{\pm0.82}$ \\
\textbf{F}: E + Replay memory & 79.8$_{\pm0.03}$ & 82.8$_{\pm1.09}$ \\

\hline

\textbf{AR-TTA (Ours)} with random memory selection & 77.1$_{\pm0.36}$ & 83.7$_{\pm0.81}$ \\

\textbf{AR-TTA (Ours)} & 78.8$_{\pm0.13}$ & 83.7$_{\pm0.64}$ \\

\hline

\end{tabular}}
\end{table}

\vspace{-0.5em}
\subsubsection{Dynamic Batch Norm Statistics Analysis}

\begin{wrapfigure}{r}{0.5\textwidth}
  \begin{center}
    \vspace{-2.2em}
    \includegraphics[trim={0 0 1cm 1.3cm},clip, width=0.4\columnwidth]{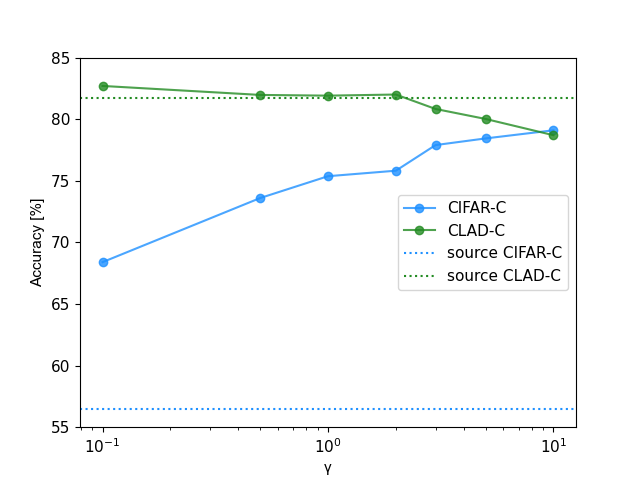}
  \end{center}
  \vspace{-0.5em}
  \caption{The relationship between mean classification accuracy (\%) and the value of parameter $\gamma$ for CIFAR10C and CLAD-C benchmarks.}
  \label{fig:bn_dist_scale}
  \vspace{-1em}
\end{wrapfigure}

The $\gamma$ is a scale parameter of the distance between distributions $D(\phi^S, \phi_t^T)$. It determines the magnitude of the calculated values of $\beta$, which is used for linear interpolation between the saved source batch normalization (BN) statistics $\phi^S$ and the BN statistics calculated from the current batch $\phi_t^T$. The higher the value of $\gamma$, the higher the values of $\beta$ tend to be. At the same time, the higher the $\beta$ values, the more influence BN statistics from current batch have on interpolation and calculation of the finally used BN statistics. In Figure~\ref{fig:bn_dist_scale} we show the relationship between $\gamma$ parameter value and mean accuracy of our AR-TTA method for CIFAR10-to-CIFAR10C and CLAD-C benchmarks. We can see the contradicting trend between the two benchmarks. This suggests that the discrepancy in the data distribution between the source domain and the estimated distribution for each test data batch is more prominent in CIFAR10C compared to CLAD-C. This is in agreement with the results of the BN-1~\cite{bn_stats_adapt} baseline method. BN-1 discards the BN statistics from the source data. Its performance was significantly better on CIFAR10C and worse on CLAD-C, compared to the fixed source model. 

\vspace{-0.5em}

\subsection{Implementation Details}
\label{sec3}
The results are averaged between 3 random seeds. Samples from CIFAR10C and \mbox{ImageNet-C} are shuffled. Considering the sequential nature of data in CLAD-C and \mbox{SHIFT-C} benchmarks (video sequences), we did not want to shuffle images. Instead, we trained 3 source models with 3 different seeds and averaged the results between experiments with different models.
We test the method in a continual manner on every benchmark, which means that the methods continually adapt the models without the reset to the source state in between the domains.

Implementations of the compared methods were taken from their official code repositories. We use all hyper-parameters and optimizers suggested by the papers or found in the code. We follow the standard model architectures used in TTA experiments and use WideResnet28 for CIFAR10C and CIFAR10.1, and ResNet50 for ImageNet-C, CLAD-C, and SHIFT-C. Moreover, since we use a smaller batch size (BS) of 10 and benchmarks that have not been used before in TTA, we search for the optimal learning rate (LR) for each method. We focus on lowering the LR, considering the decreased batch size. Additionally, we search for the $\epsilon$ hyperparameter of EATA to correctly reject samples for adaptation. The results of the parameter search can be found in Table~\ref{tab:grid_search}. The details and parameters used for each method are described below.

\paragraph{TENT~\cite{TENT}} We use Adam optimizer with LR = 0.00025 for CIFAR10.1 and LR = 0.00003125 for every other tested dataset. In the original paper, TENT uses LR = 0.001 for all the datasets except ImageNet, but it performed worse with this value in our setup.

\paragraph{CoTTA~\cite{COTTA}} Adam optimizer with LR = 0.00025 is used for every tested benchmark, except ImageNet-C for which LR was equal to 0.00003125. The original implementation set LR to 0.001, but with an adjusted value, it achieved better results. We follow the suggestions for other hyperparameter values given by the authors. The restoration probability $p$ is set to 0.01, the smoothing factor of the exponential moving average of teacher weights $\alpha$ is equal to 0.999, and the confidence threshold for applying augmentations $p_{th}$ is set to 0.92.

\paragraph{EATA~\cite{eata}} We use the SGD optimizer with a momentum of 0.9 and LR of 0.00025 for CIFAR10C, ImageNet-C, and CLAD-C. LR for SHIFT-C is equal to 0.00003125 and 0.001 for CIFAR10.1. The original EATA paper uses an LR value of 0.005/0.00025 for CIFAR10C/ImageNet-C, but they used BS = 64. After the search for the optimal $\epsilon$ parameter value for filtering redundant samples, we set it to 0.05/0.6 for CLAD-C/SHIFT. The value of $\epsilon$ for CIFAR10C and CIFAR10.1/ImageNet-C is equal to 0.4/0.05, as in the original paper. The entropy constant $E_0$ is set to the standard value of 0.4$\times \ln C$, where $C$ was the number of classes, following the original paper and~\cite{SAR}. The trade-off parameter $\beta$ is equal to 1, and 2000 samples are used to calculate the fisher importance of model weights as for the CIFAR10 dataset in the original paper.

\paragraph{SAR~\cite{SAR}} SGD optimizer is used with the momentum of 0.9 and LR = 0.001 for both CIFAR10C and CIFAR10.1, and LR = 0.00025 for ImageNet-C, CLAD-C, and SHIFT-C. It almost aligns with the authors' choice since, in original experiments, they used a learning rate equal to 0.00025/0.001 for ResNet/Vit models. The parameter $E_0$ is set to 0.4$\times \ln C$, as in the paper, similarly to EATA. We follow the authors' choice of a constant reset threshold value $e_0$ of 0.2, and a moving average factor equal to 0.9. The radius parameter $\rho$ is set to the default value of 0.05.

\paragraph{RMT~\cite{dobler2023CVPRmeanteacher}} Adam optimizer is used with LR = 0.00025 for CIFAR10C, CLAD-C, and SHIFT-C. LR is equal to 0.00003125  for CIFAR10.1 and ImageNet-C, all following the grid search. Following the original implementation, we use temperature $\tau$ for contrastive loss set to 0.1 and momentum param $\alpha$ utilized to update the mean teacher equal to 0.999.

\paragraph{AR-TTA (Ours)} We use an SGD optimizer with momentum of 0.9. We set LR of 0.00025 for both ImageNet-C and CIFAR10.1, and 0.001 for the rest of the benchmarks. The scale hyper-parameter $\gamma$ is set to 0.1 for CIFAR10.1, CLAD-C, and SHIFT-C. It is equal to 10 for ImageNet-C and CIFAR10C. $\alpha$ value for weighting the $\beta_{ema}$ is equal to 0.2. We set the initial $\beta_{ema}$ value to 0.1. The $\psi$ and $\rho$ parameters used for beta distribution to sample $\lambda$ for mixup is equal to the standard value of 0.4. We store 2000 of exemplars from source data for memory replay.

\begin{table*}[h!]
\centering
\caption{Mean classification accuracy~(\%) for CIFAR10C, ImageNet-C, CIFAR10.1, CLAD-C, and SHIFT-C continual test-time adaptation task for compared state-of-the-art methods with different learning rates and EATA's $\epsilon$ parameter.}
\label{tab:grid_search}
\vspace{2mm}
\scalebox{0.7}{
\begin{tabular}{l|c|c|c|c|c|c|c}
\hline

Method & learning rate & $\epsilon$ & CIFAR10C & CIFAR10.1 & CLAD-C & SHIFT-C & ImageNet-C \\ \hline

\multirow{3}{*}{CoTTA~\cite{COTTA}} 
& 0.001 & - & 49.3 & 79.3 & 71.5 & 74.3 & 3.8 \\
& 0.00025 & - & 75.7 & 82.3 & 71.8 & 78.6 & 10.6 \\
& 0.00003125 & - & 74.5 & 81.8 & 71.8 & 76.2 & 15.3 \\
\hline

\multirow{3}{*}{TENT~\cite{TENT}} 
& 0.001 & - & 24.3 & 81.2 & 64.4 & 63.4 & 0.6 \\
& 0.00025 & - & 72.3 & 82.3 & 71.0 & 75.3 & 3.1 \\
& 0.00003125 & - & 76.7 & 81.4 & 71.1 & 82.7 & 29.3 \\
\hline

\multirow{3}{*}{SAR~\cite{SAR}} 
& 0.001 & - & 75.2 & 81.3 & 70.6 & 86.0 & 11.3 \\
& 0.00025 & - & 75.1 & 81.3 & 70.6 & 86.0 & 31.5 \\
& 0.00003125 & - & 75.0 & 81.3 & 70.6 & 86.0 & 28.8 \\
\hline

\multirow{3}{*}{RMT~\cite{dobler2023CVPRmeanteacher}} 
& 0.001 & - & 83.0 & 81.1 & 76.0 & 93.1 & 30.2 \\
& 0.00025 & - & 83.1 & 81.9 & 75.3 & 95.9 & 28.6 \\
& 0.00003125 & - & 81.5 & 83.3 & 74.5 & 93.1 & 30.5 \\
\hline

\multirow{12}{*}{EATA~\cite{eata}} 
& 0.001 & 0.60 & - & 82.6 & 70.1 & 80.4 & - \\
& 0.001 & 0.40 & 76.3 & 82.9 & 70.6 & 80.4 & - \\
& 0.001 & 0.10 & - & 82.4 & 70.6 & 86.0 & - \\
& 0.001 & 0.05 & - & 82.4 & 70.6 & 86.0 & 27.3 \\

& 0.00025 & 0.60 & - & 82.4 & 70.5 & 85.6 & - \\
& 0.00025 & 0.40 & 78.2 & 82.6 & 70.6 & 86.1 & - \\
& 0.00025 & 0.10 & - & 82.4 & 70.6  & 86.0 & - \\
& 0.00025 & 0.05 & - & 82.4 & 70.7 & 86.0 & 31.7 \\

& 0.00003125 & 0.60 & - & 82.4 & 70.6 & 86.1 & - \\
& 0.00003125 & 0.40 & 76.5 & 82.4 & 70.6 & 86.0 & - \\
& 0.00003125 & 0.10 & - & 82.4 & 70.6  & 86.0 & - \\
& 0.00003125 & 0.05 & - & 82.4 & 70.6 & 86.0 & 31.6 \\
\hline
\end{tabular}}
\end{table*}

\subsection{Benchmarks}
\label{sec4}

\subsubsection{CIFAR10C And ImageNet-C Benchmarks Details}
CIFAR10C and ImageNet-C are widely used datasets in TTA. They involve training the source model on train split of clean CIFAR10/ImageNet datasets~\cite{cifar, imagenet} and test-time adaptation on CIFAR10C/ImageNet-C. CIFAR10C and ImageNet-C consist of images from clean datasets which were modified by 15 types of corruptions with 5 levels of severity\cite{hendrycks2019robustness}. They were first used for evaluating the robustness of neural network models and are now widely utilized for testing the adaptation capabilities of TTA methods. We test the adaptation on a standard sequence of the highest corruption severity level 5, frequently utilized by previous approaches~\cite{COTTA, eata, dobler2023CVPRmeanteacher}. For ImageNet-C we utilize a subset of 5000 samples for each corruption, based on \textit{RobustBench} library~\cite{croce2020robustbench}, following~\cite{COTTA}.

\subsubsection{CIFAR10.1 Benchmark Details}
CIFAR10.1~\cite{cifar10_1} was designed to minimize the distribution shift relative to the original CIFAR10 dataset~\cite{cifar}. It contains roughly 2000 test images. The images in CIFAR10.1 are a~subset of the TinyImages dataset~\cite{tinyimages}. The source model utilized for testing on this benchmark was pre-trained on the original CIFAR10 dataset.

\subsubsection{CLAD-C Benchmark Details}
CLAD-C~\cite{verwimp2023clad} is an online classification benchmark for autonomous driving with the goal of introducing a more realistic testing bed for continual learning. It consists of natural, temporal correlated, and continuous distribution shifts created by utilizing the data from SODA10M dataset~\cite{han2021soda10m}. The images taken at different locations, times of day, and weather, are chronologically ordered, inducing distribution shifts in labels and domains. 

The classification task was created by cutting out the annotated 2D bounding boxes of six classes and using them as separate images for classification. Bounding boxes with fewer than 1024 pixels were discarded. The images are padded by their shortest axis (modify the aspect ratio to 1:1) and resized to 32x32. For the ResNet50 model, we additionally resize them to 224x224. 

Since it is designed for testing the continual learning setup and the model is originally supposed to be trained sequentially on the train sequences, we slightly modify the setup and pre-train the source model on the first train sequence. TTA is continually tested on the 5 remaining ones with a total number of 17092 images.

\subsubsection{SHIFT-C Benchmark Details}
The SHIFT-C benchmark is created using the SHIFT dataset~\cite{shift2022}. It consists of multiple types of autonomous driving data from the CARLA Simulator~\cite{carla}. We used RGB images from the front view of a car with discrete domain shifts and bounding box annotations. More specifically, we download the required data with the script from SHIFT's website \url{https://www.vis.xyz/shift/}, using the following command:

\noindent\begin{minipage}{\textwidth}
\begin{lstlisting}
python download.py --view "front" \
--group "[img, det_2d]" \
--split "[train, val]" \
--framerate "images" \
--shift "discrete" TARGET_DIR
\end{lstlisting}
\end{minipage}
To load the data for experiments, we utilized \textit{shift-dev} repository: \url{https://github.com/SysCV/shift-dev}.

\begin{figure*}[hbt!]
    \centering
    \includegraphics[width=\textwidth]{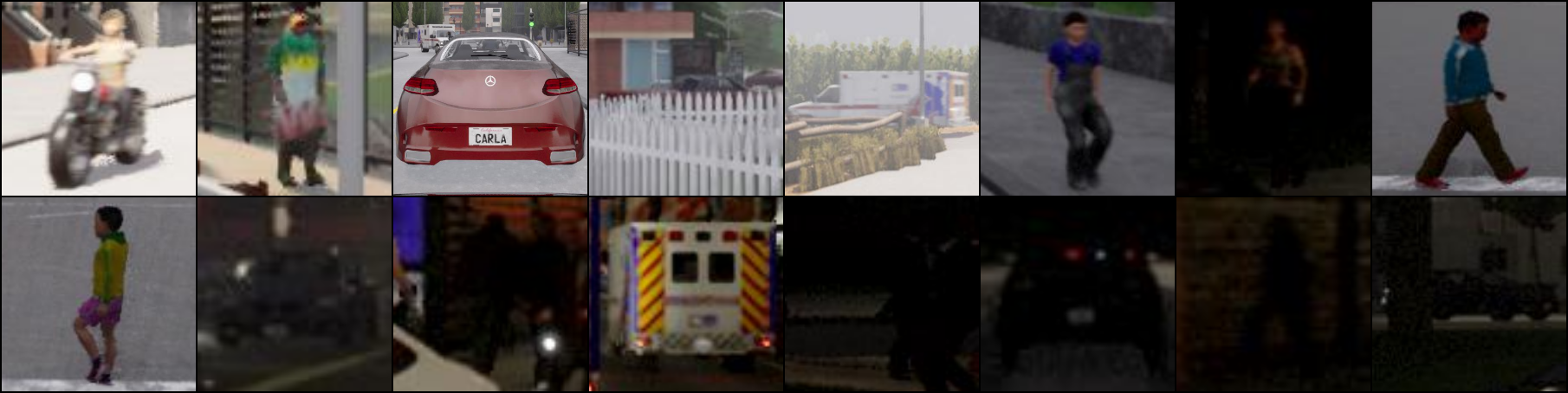}
\vspace{2mm}
    \caption{Example images sourced from various domains within the SHIFT-C benchmark.}
    \label{fig:shift_images}
\end{figure*}

\begin{figure*}[hbt!]
    \centering
    \includegraphics[width=0.7\textwidth]{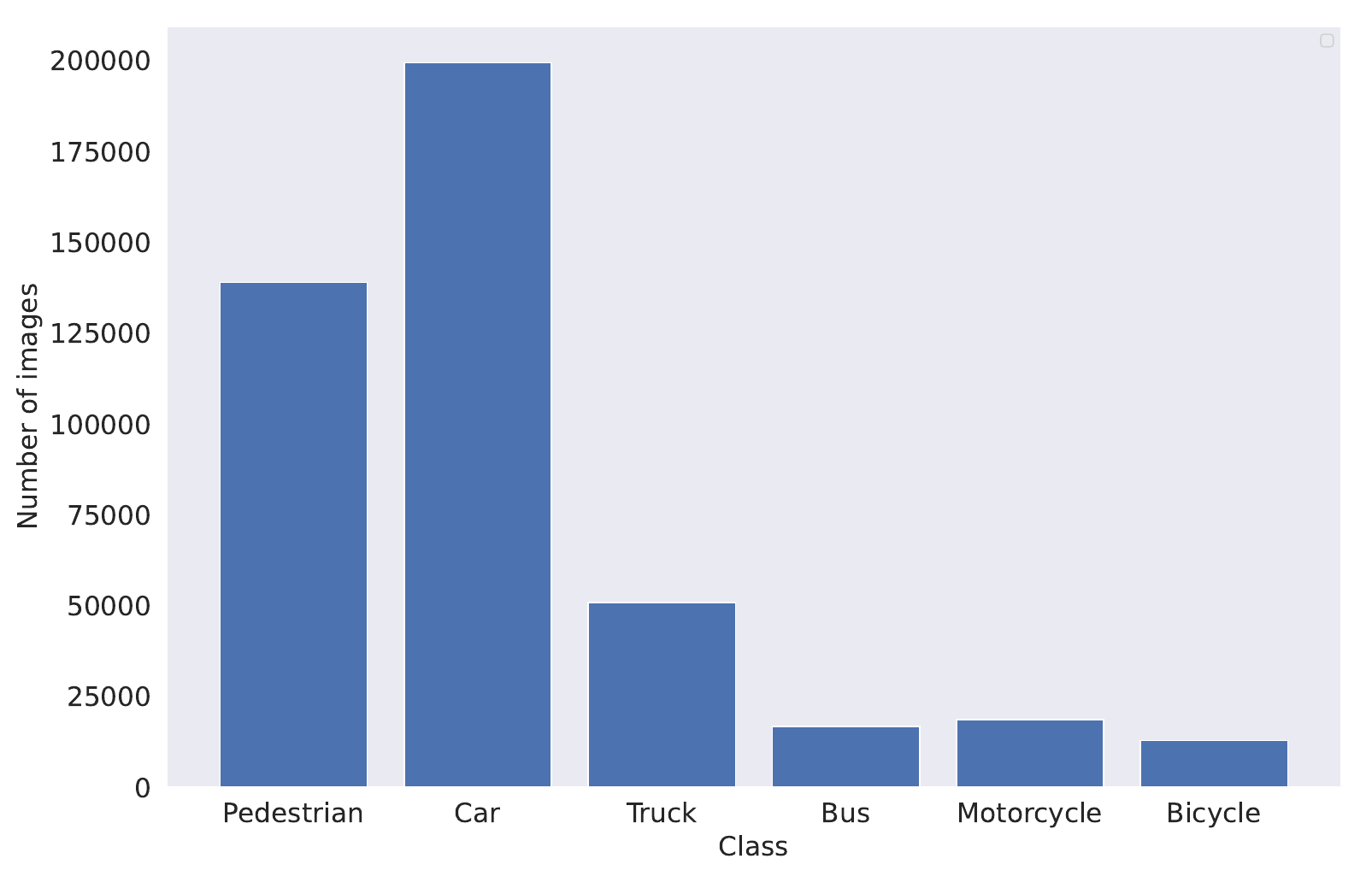}
    \vspace{2mm}
    \caption{SHIFT-C benchmark class distribution.}
    \label{fig:shift_class_distrib}
\end{figure*}

We create an image classification task data, following the steps from the CLAD-C\cite{verwimp2023clad} benchmark. Bounding boxes in the dataset are categorized into six classes, and so are the created images. Example images are displayed in Figure~\ref{fig:shift_images}. We present a class distribution in Figure~\ref{fig:shift_class_distrib}. 

We distinguish between domains by the course annotations of time of day and weather. The source model is trained on images from train split, taken in the daytime in clear weather. The TTA is also tested on data from the train split but from different weather conditions and times of the day. Details about the size of each domain can be found in Table~\ref{tab:shift_dataset_details}.

\begin{table*}[h!]
\centering
\caption{The number of samples in each domain in SHIFT-C benchmark}\label{tab:shift_dataset_details}
\vspace{2mm}
\scalebox{0.9}{
\begin{tabular}{c|c|c|c}
\hline

\textbf{Domain nr} & \textbf{Time of day} & \textbf{Weather} & \textbf{Number of images} \\
\hline
Source data &  \multirow[c]{5}{*}{daytime} & clean & 57039 \\
1 &  & cloudy & 41253 \\
2 &  & overcast & 20497 \\
3 &  & rainy & 59457 \\
4 &  & foggy & 38590 \\

\hline

5 & \multirow[c]{5}{*}{dawn/dusk} & clear & 29543 \\
6 &  & cloudy & 19985 \\
7 &  & overcast & 9901 \\
8 &  & rainy & 26677 \\
9 &  & foggy & 20258 \\

\hline

10 & \multirow[c]{5}{*}{night} & clear & 28639 \\
11 &  & cloudy & 18068\\
12 &  & overcast & 9471 \\
13 &  & rainy & 32864\\
14 &  & foggy & 25464 \\

\hline

\multicolumn{3}{l}{Sum} & 437706 \\

\hline
\end{tabular}}
\end{table*}

\subsection{Additional Results}
\label{sec2}
\begin{figure*}[ht!]
    \centering
    \includegraphics[width=0.9\textwidth]{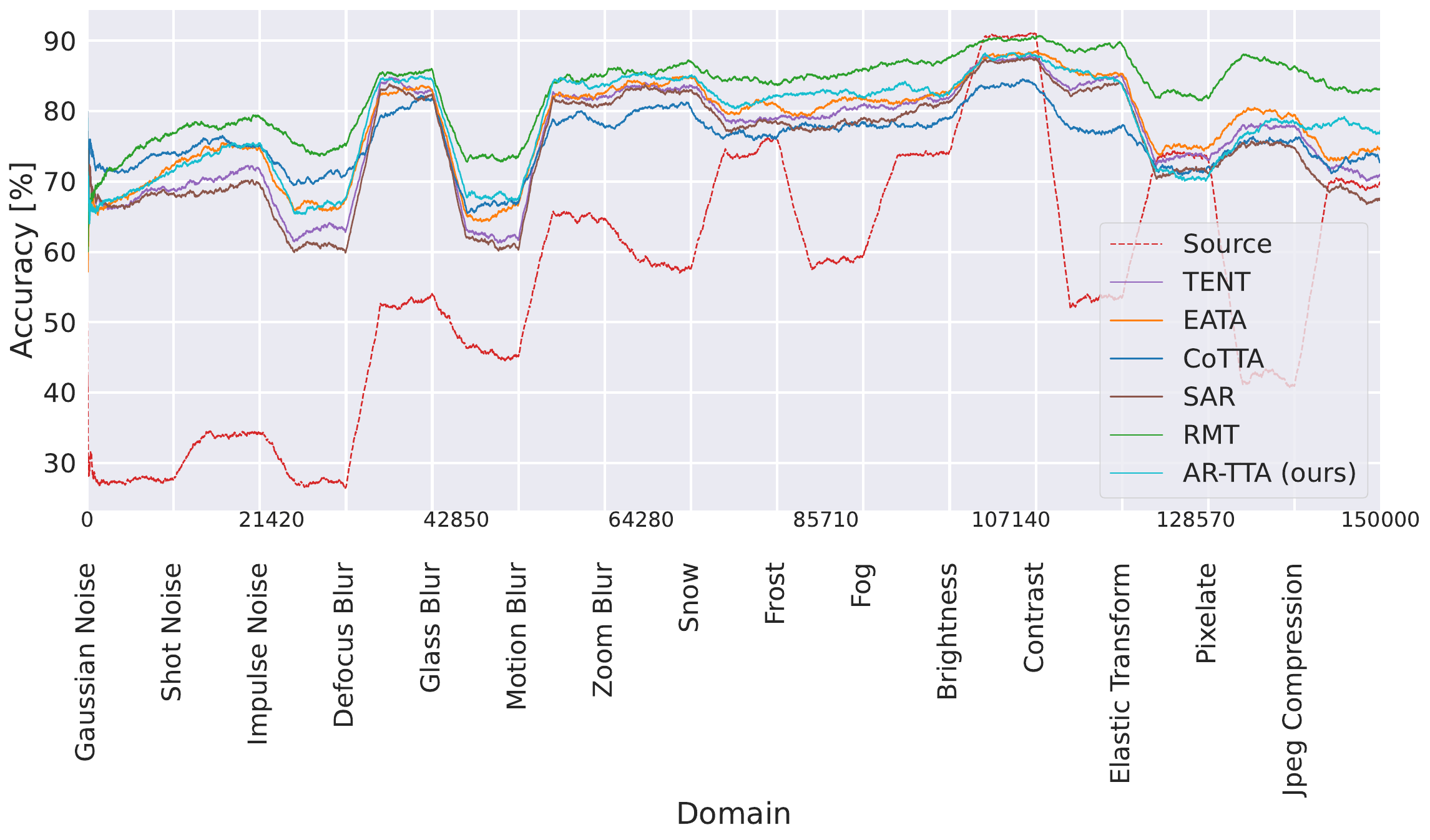}
    \vspace{2mm}
    \caption{Batch-wise classification accuracy (\%) averaged in a window of 400 batches on CIFAR10C benchmark for the chosen methods continually adapted to the sequences of data. The major ticks on the x-axis symbolize the beginning of the next sequence and, at the same time, a different domain. Minor ticks on the x-axis (numbers) indicate the image number. Best viewed in color.}
    \label{fig:cifar_batchwise}
\end{figure*}

\begin{figure*}[ht!]
    \centering
    \includegraphics[width=0.9\textwidth]{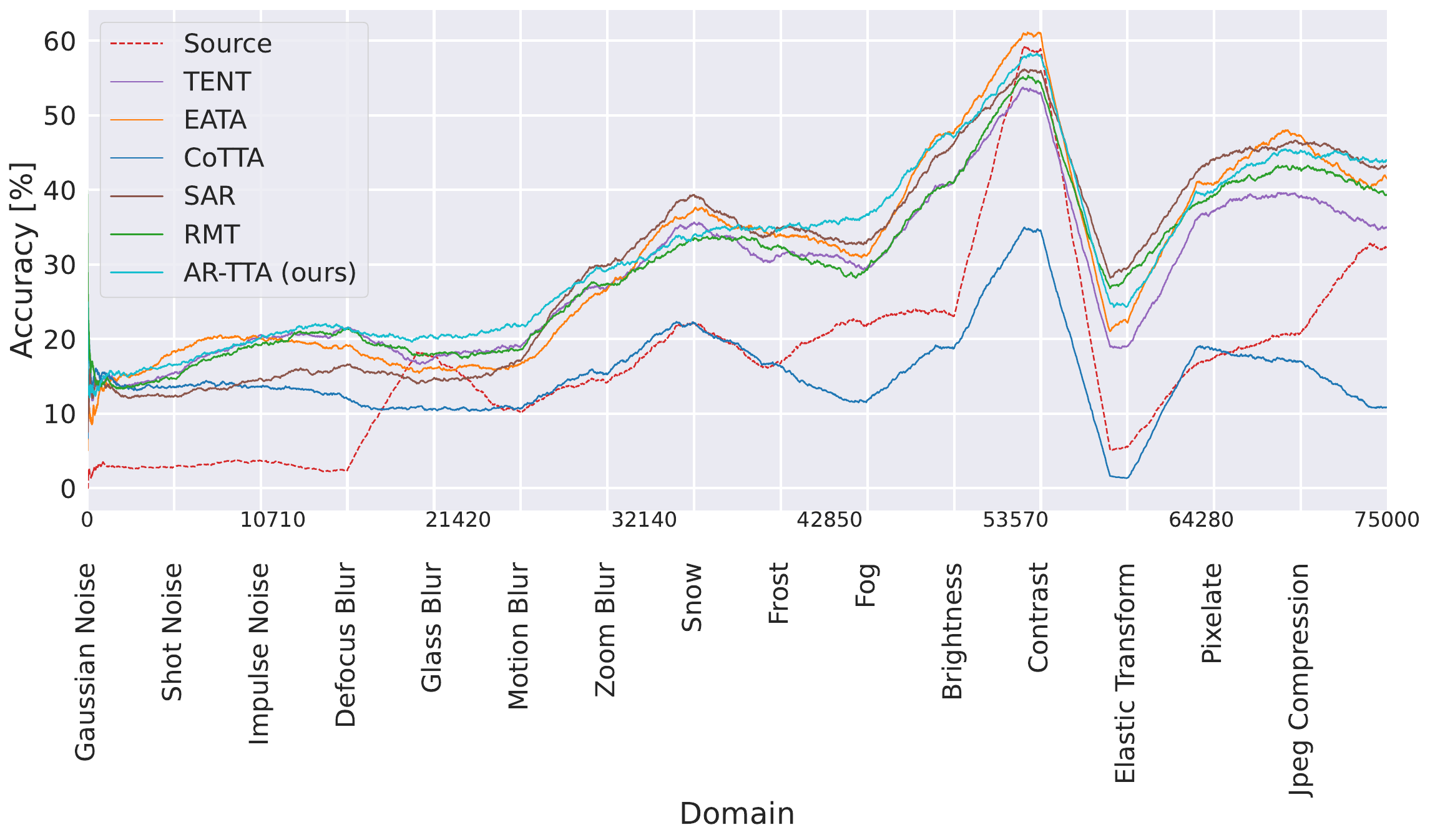}
    \vspace{2mm}
    \caption{Batch-wise classification accuracy (\%) averaged in a window of 400 batches on ImageNet-C benchmark for the chosen methods continually adapted to the sequences of data. The major ticks on the x-axis symbolize the beginning of the next sequence and, at the same time, a different domain. Minor ticks on the x-axis (numbers) indicate the image number. Best viewed in color.}
    \label{fig:imagenet_batchwise}
\end{figure*}

\begin{figure*}[ht!]
    \centering
    \includegraphics[width=0.9\textwidth]{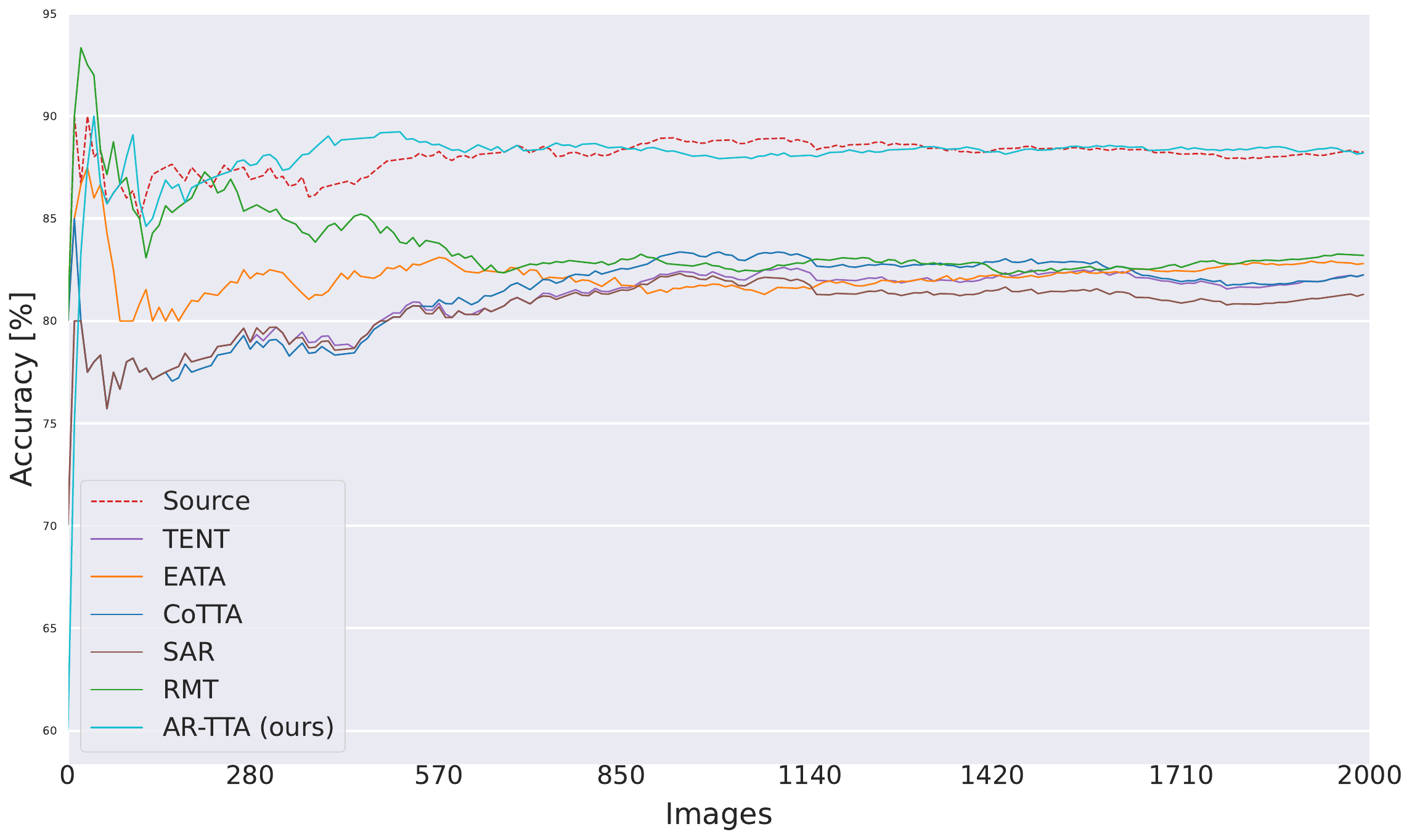}
    \vspace{2mm}
    \caption{Batch-wise classification accuracy (\%) averaged in a window of 400 batches on CIFAR10.1 benchmark for the chosen methods continually adapted to the sequences of data. Best viewed in color.}
    \label{fig:cifar10.1}
\end{figure*}

\begin{figure*}[ht!]
    \centering
    \includegraphics[width=0.95\textwidth]{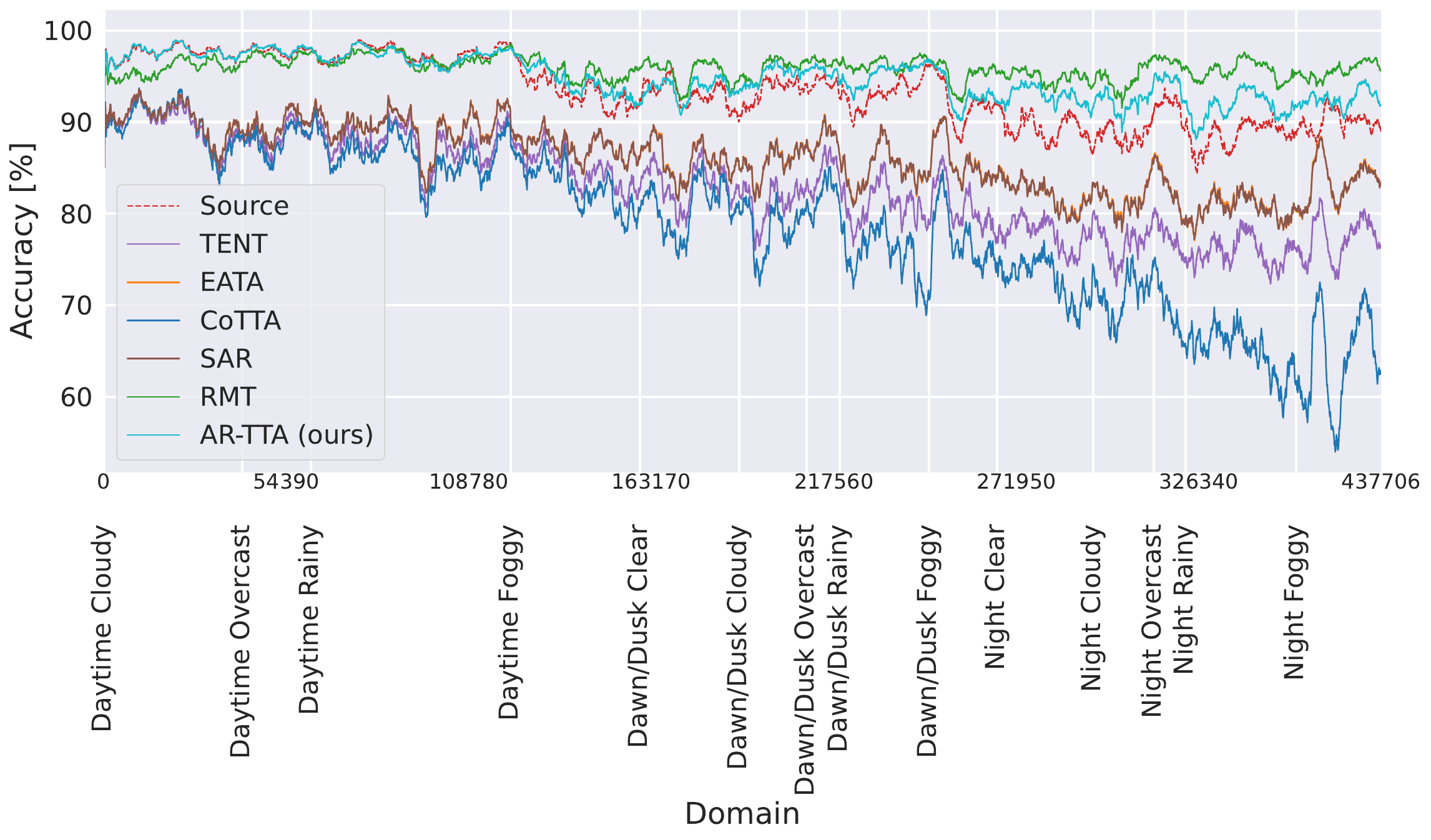}
    \vspace{2mm}
    \caption{Batch-wise classification accuracy (\%) averaged in a window of 500 batches on SHIFT-C benchmark for the chosen methods continually adapted to the sequences of data. The major ticks on the x-axis symbolize the beginning of the next sequence and, at the same time, a different domain. Minor ticks on the x-axis (numbers) indicate the image number. Best viewed in color.}
    \label{fig:shift_batchwise}
\end{figure*}

We present batch-wise accuracy plots for the CIFAR10C, ImageNet-C, CIFAR10.1, and SHIFT-C benchmarks in Figures~\ref{fig:cifar_batchwise},~\ref{fig:imagenet_batchwise},~\ref{fig:cifar10.1},~\ref{fig:shift_batchwise}, respectively. Moreover, the full results on \mbox{CIFAR10C} can be found in Table~\ref{tab:cifar10}.

\begin{table*}[ht!]
\centering
\caption{Classification accuracy~(\%) for the standard CIFAR10C online continual test-time adaptation task.}
    \vspace{2mm}
\label{tab:cifar10}
\scalebox{0.57}{
\begin{tabular}{l|ccccccccccccccc|c}
\multicolumn{1}{l}{}& \multicolumn{15}{c}{ $t\xrightarrow{\hspace*{14cm}}$}& \\ \hline
Method & \rotatebox[origin=c]{70}{Gaussian} & \rotatebox[origin=c]{70}{shot} & \rotatebox[origin=c]{70}{impulse} & \rotatebox[origin=c]{70}{defocus} & \rotatebox[origin=c]{70}{glass} & \rotatebox[origin=c]{70}{motion} & \rotatebox[origin=c]{70}{zoom} & \rotatebox[origin=c]{70}{snow} & \rotatebox[origin=c]{70}{frost} & \rotatebox[origin=c]{70}{fog}  & \rotatebox[origin=c]{70}{brightness} & \rotatebox[origin=c]{70}{contrast} & \rotatebox[origin=c]{70}{elastic\_trans} & \rotatebox[origin=c]{70}{pixelate} & \rotatebox[origin=c]{70}{jpeg} & Mean \\ \hline

Source & 27.7 & 34.3 & 27.1 & 53.1 & 45.7 & 65.2 & 58.0 & 74.9 & 58.7 & 74.0 & 90.7 & 53.3 & 73.4 & 41.6 & 69.7 & 56.5 \\

BN-1 & 67.3 & 69.4 & 59.7 & 82.7 & 60.4 & 81.4 & 83.0 & 78.1 & 77.7 & 80.6 & 87.3 & 83.4 & 71.4 & 75.3 & 67.9 & 75.0 \\

TENT~\cite{TENT} & 67.9 & 71.4 & 62.5 & 83.2 & 62.9 & 82.1 & 83.8 & 79.5 & 79.7 & 81.4 & 87.8 & 84.3 & 73.5 & 78.2 & 71.6 & 76.7 \\

EATA~\cite{eata} & 70.3 & 74.9 & 67.1 & 83.0 & 65.6 & 82.3 & 84.0 & 80.3 & 81.4 & 82.2 & 88.0 & 85.1 & 74.7 & 80.1 & 73.8 & 78.2 \\

CoTTA~\cite{COTTA} & \textbf{72.5} & \textbf{76.4} & \textbf{70.5} & 80.6 & 66.6 & 78.3 & 80.1 & 75.8 & 77.0 & 77.1 & 83.8 & 77.3 & 72.0 & 75.5 & 72.2 & 75.7 \\

SAR~\cite{SAR} & 67.4 & 69.6 & 60.8 & 82.6 & 61.4 & 81.5 & 82.8 & 78.1 & 77.7 & 80.5 & 87.4 & 83.4 & 71.5 & 75.2 & 68.2 & 75.2 \\

RMT w/o replay~\cite{dobler2023CVPRmeanteacher} & 73.2 & 77.5 & 72.1 & 79.6 & 69.8 & 78.2 & 79.7 & 76.9 & 78.0 & 79.9 & 83.0 & 81.6 & 76.5 & 80.4 & 77.3 & 77.6 \\
RMT~\cite{dobler2023CVPRmeanteacher} & 74.8 & 78.6 & 74.8 & \textbf{85.5} & \textbf{73.1} & \textbf{84.7} & \textbf{86.1} & \textbf{84.2} & \textbf{85.4} & \textbf{87.2} & \textbf{90.1} & \textbf{89.1} & \textbf{82.2} & \textbf{87.1} & \textbf{83.2} & \textbf{83.1} \\

\hline

AR-TTA (Ours) w/o replay & 69.5 & 73.6 & 63.3 & 83.5 & 63.0 & 82.5 & 84.5 & 80.2 & 80.4 & 81.9 & \textbf{88.4} & 83.8 & 74.2 & 76.9 & 74.5 & 77.3$_{\pm0.07}$ \\

AR-TTA (Ours) & 69.2 & 74.8 & 66.4 & 84.5 & 67.8 & 83.7 & 85.2 & 81.4 & 82.7 & 83.4 & 88.0 & 84.7 & 73.9 & 78.6 & 77.0 & 78.8$_{\pm0.13}$ \\

\hline
\end{tabular}}
\end{table*}

\end{document}